\documentclass[english]{article}
\usepackage[T1]{fontenc}
\usepackage[latin9]{inputenc}
\usepackage{amsmath}
\usepackage{graphicx}

\makeatletter

\providecommand{\tabularnewline}{\\}

\usepackage{ijcai17}

\usepackage[english]{babel}
\usepackage{amsbsy}

\usepackage{microtype}
\usepackage{flushend}
\usepackage{times}
\usepackage{helvet}
\usepackage{courier}
\usepackage{algorithmic}

\makeatother

\usepackage{babel}
\begin{document}

\title{Graph Classification via Deep Learning with Virtual Nodes}

\author{Trang Pham$^{\natural}$, Truyen Tran$^{\natural}$, Hoa Dam$^{\flat}$,
Svetha Venkatesh$^{\natural}$\\
$^{\natural}$Centre for Pattern Recognition and Data Analytics\\
Deakin University, Geelong, Australia\\
\{\textit{phtra, truyen.tran, svetha.venkatesh}\}\textit{@deakin.edu.au;}\\
$^{\flat}$School of Computing and Information Technology\\
University of Wollongong, Australia\\
\textit{hoa@uow.edu.au}}
\maketitle
\begin{abstract}
Learning representation for graph classification turns a variable-size
graph into a fixed-size vector (or matrix). Such a representation
works nicely with algebraic manipulations. Here we introduce a simple
method to augment an attributed graph with a virtual node that is
bidirectionally connected to all existing nodes. The virtual node
represents the latent aspects of the graph, which are not immediately
available from the attributes and local connectivity structures. The
expanded graph is then put through any node representation method.
The representation of the virtual node is then the representation
of the entire graph. In this paper, we use the recently introduced
Column Network for the expanded graph, resulting in a new end-to-end
graph classification model dubbed Virtual Column Network (VCN). The
model is validated on two tasks: (i) predicting bio-activity of chemical
compounds, and (ii) finding software vulnerability from source code.
Results demonstrate that VCN is competitive against well-established
rivals.

\end{abstract}
\global\long\def\xb{\boldsymbol{x}}
\global\long\def\yb{\boldsymbol{y}}
\global\long\def\eb{\boldsymbol{e}}
\global\long\def\zb{\boldsymbol{z}}
\global\long\def\hb{\boldsymbol{h}}
\global\long\def\ab{\boldsymbol{a}}
\global\long\def\bb{\boldsymbol{b}}
\global\long\def\cb{\boldsymbol{c}}
\global\long\def\sigmab{\boldsymbol{\sigma}}
\global\long\def\gammab{\boldsymbol{\gamma}}
\global\long\def\alphab{\boldsymbol{\alpha}}
\global\long\def\rb{\boldsymbol{r}}
\global\long\def\gb{\boldsymbol{g}}
\global\long\def\Deltab{\boldsymbol{\Delta}}
\global\long\def\wb{\boldsymbol{w}}
\global\long\def\vb{\boldsymbol{v}}
\global\long\def\eb{\boldsymbol{e}}
\global\long\def\sb{\boldsymbol{s}}
\global\long\def\ub{\boldsymbol{u}}

\section{Intro}

Deep learning has achieved record-breaking successes in domains with
regular grid (e.g., via CNN) or chain-like (e.g., via RNN) structures
\cite{lecun2015deep}. However, many, if not most, real-world structured
problems are best modelled using graphs with irregular connectivity.
These include, for example, chemical compounds, proteins, RNAs, function
calls in software, brain activity networks, and social networks. A
canonical task is \emph{graph classification}, that is, assigning
class labels to a graph instance (e.g., chemical compound as its activity
against cancer cells).

We study a generic class known as attributed graphs whose nodes can
have attributes, and edges can be multi-typed and directed. We aim
to efficiently \emph{learn distributed representation of graph}, that
is, a map that turns variable-size graphs into fixed-size vectors
or matrices. Such a representation would benefit greatly from a powerful
pool of data manipulation tools. This rules out traditional approaches
such as graph kernels \cite{vishwanathan2010graph} and graph feature
engineering \cite{choetkiertikul2017predicting-tse}, which could
be either computation or labor intensive.

Several recent neural networks defined on graphs, such as Graph Neural
Network \cite{scarselli2009graph}, diffusion-CNN \cite{atwood2016diffusion},
and Column Network \cite{pham2017column}, start with node representations
by taking into account of the neighborhood structures, typically through
convolution and/or recurrent operations. Node representations can
then be aggregated into graph representation. It is akin to representing
a document\footnote{A document can be considered as a linear graph of words.}
by first embedding words into vectors (e.g., through word2vec) then
combining them (e.g., by weighted averaging using attention). We conjecture
that a better way is to learn graph representation directly and simultaneously
with node representation\footnote{This is akin to the spirit of paragraph2vec \cite{le2014distributed}.}.

Our main idea is to augment the original graph with a \emph{virtual
node} to represent the latent aspects of the graph. The virtual node
is bidirectionally connected to all existing real nodes. The virtual
node assumes either empty attributes or auxiliary information which
are not readily available in the original graph. The augmented-graph
is passed through the existing graph neural networks for computing
node representation. The graph representation is then a vector representation
of the virtual node. 

For concreteness, we materialize the idea of virtual node using Column
Network \cite{pham2017column}, an architecture for node classification.
With a differentiable classifier (e.g., a feedforward net), the network
is an end-to-end solution for graph classification, which we name
\emph{Virtual Column Network} (VCN). We validate the VCN on two applications:
predicting bio-activity of chemical compounds, and assessing vulnerability
in software code and the results are promising.

\section{Models}

\begin{figure}
\centering{}\includegraphics[bb=60bp 135bp 650bp 490bp,clip,width=0.98\columnwidth]{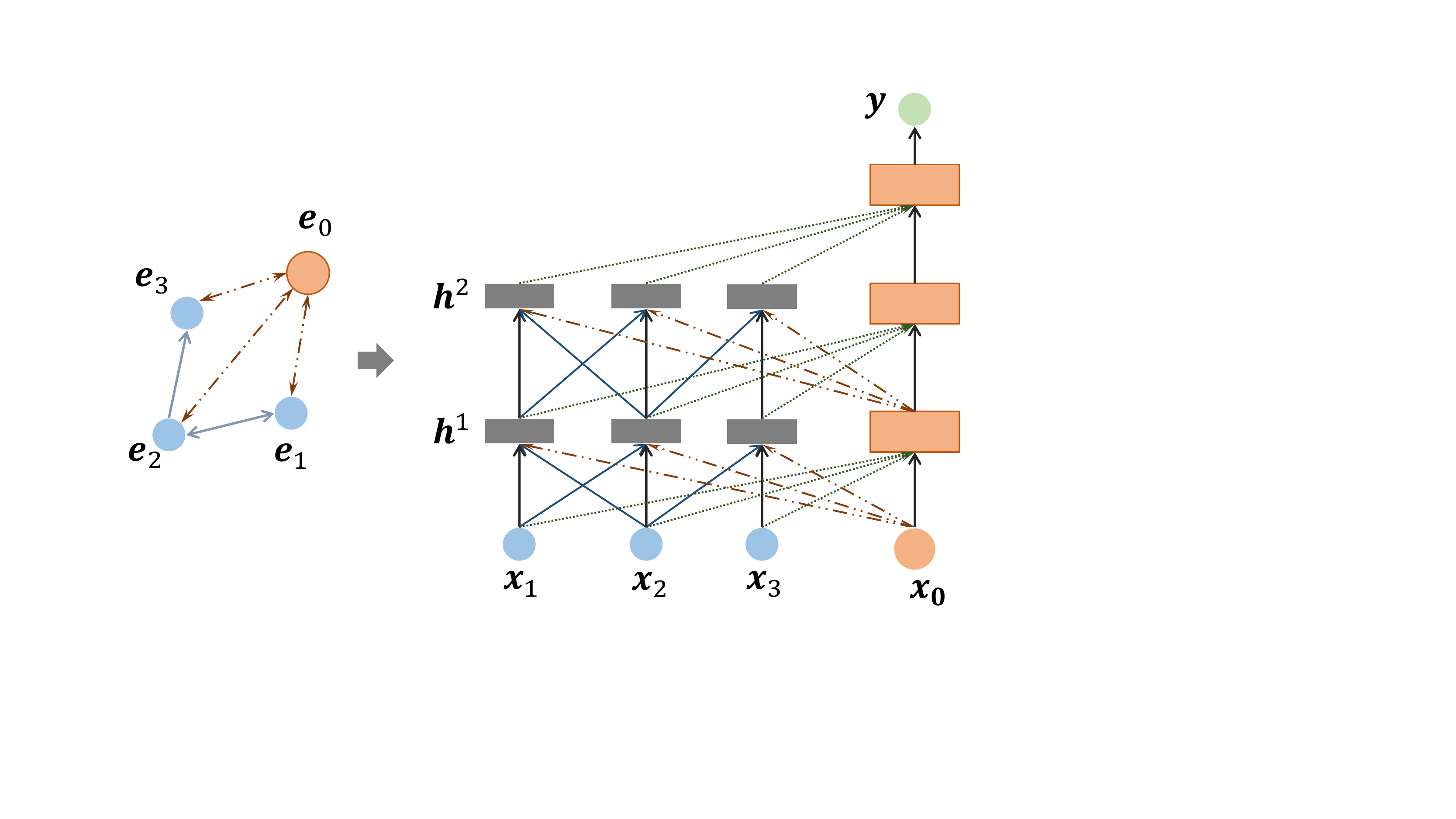}\caption{Virtual Column Network (VCN) = Column Network + Virtual Node. (Left)
A graph of 3 nodes augmented with a virtual node $\protect\eb_{0}$
connecting to all nodes. (Right) The VCN model for the graph, where
$\protect\xb_{0}$ is vector of graph descriptors (if any), $\protect\xb_{1},\protect\xb_{2},\protect\xb_{3}$
are node attributes, and $y$ is the graph label. Boxes are hidden
layers. \label{fig:Virtual-Column-Network}}
\end{figure}

In this section we present Virtual Column Network (VCN), a realization
of the idea of virtual node for graph classification using a recent
node representation method known as Column Network (CLN) \cite{pham2017column}.
VCN is applicable to graphs of multi-typed edges and attributed nodes.

\subsection{Column Network}

Given a attributed graph, Column Network is a recurrent neural architecture
defined on it. For each network, there are multiple interacting recurrent
nets (called \emph{columns}, as analogy to cortical columns in brain
\cite{mountcastle1997columnar}), each of which is responsible to
a node. Fig.~\ref{fig:Virtual-Column-Network} illustrates the VCN
model, where the columns of $\xb_{1},\xb_{2},\xb_{3}$ create a Column
Network. Neural connections from one column to another reflect the
graph edge between the two corresponding nodes. More concretely, the
column for node $i$ ($i=1,...n$) at step $t$ is updated using information
from neighbors at step $t-1$ as follows: 

\begin{align*}
\cb_{ip}^{t} & =\frac{1}{\mid N_{p}\left(i\right)\mid}\sum_{j\in N_{p}\left(i\right)}\hb_{j}^{t-1}\\
\tilde{\hb}_{i}^{t} & =g\left(W\hb_{i}^{t-1}+\sum_{p}U_{p}\cb_{ip}^{t}\right)\\
\hb_{i}^{t} & =(1-\alphab_{i}^{t})\ast\hb_{i}^{t-1}+\alphab_{i}^{t}\ast\tilde{\hb}_{i}^{t}
\end{align*}
where:
\begin{itemize}
\item $\hb_{i}^{t}$ denotes the state of column $i$ at step $t$;
\item $N_{p}\left(i\right)$ is the neighborhood of node $i$ w.r.t. edge
type $p$;
\item $\cb_{ip}^{t}$ denotes the neighboring context w.r.t. edge type $p$;
\item $\tilde{\hb}_{i}^{t}$ is the candidate state, and $g(\cdot)$ is
nonlinear transformation (e.g., ReLU or tanh); $W_{i},U_{ip}$ are
parameters tied across steps. When columns are homogeneous, we can
have remove the sub-index $i$;
\item $\alphab_{i}^{t}\in(0,1)$ is the smoothing gate, which is parameterized
in the same way as $\tilde{\hb}_{i}^{t}$, and $g(\cdot)$ is typically
a sigmoid function.
\end{itemize}
At $t=1$, the state is assigned to the projection of attributes at
node $i$, i.e., $\hb_{i}^{1}=P_{i}\xb_{i}$.

\textbf{Remark}: $\cb_{i}^{t}$ plays the role of usual input at step
$t$ for RNN. Without this, the model becomes the Highway Network
\cite{srivastava2015training} with parameters typing \cite{pham2016faster}.
This also suggests a GRU alternative with a reset gate $\rb_{i}^{t}\in(0,1)$,
i.e., $\tilde{\hb}_{i}^{t}=g\left(W\left(\rb_{i}^{t}\ast\hb_{i}^{t-1}\right)+\sum_{p}U_{p}\cb_{ip}^{t}\right)$.
When there is only one neighbor per type, computing the candidate
$\tilde{\hb}_{i}^{t}$ becomes the standard convolution operation
in CNN with neighbor indexed by $p$. More specifically, $\cb_{ip}^{t}=\hb_{j}^{t-1}$
for $N_{p}\left(i\right)\equiv j$ , leading to $\tilde{\hb}_{i}^{t}=g\left(W\hb_{i}^{t-1}+\sum_{j=N_{p}\left(i\right)}U_{p}\hb_{j}^{t-1}\right)$. 

\subsection{Virtual Columns}

Column networks are compact and effective in integrating long-range
dependencies between nodes (the radius is equal to the height of the
columns). However, it was designed for node classification. For graph
classification we need a way to pool node states. A simple way is
taking an average at the end: $\bar{\hb}=\frac{1}{n}\sum_{i}\hb_{i}^{T}$.

Here we introduce a new way for integrating node states. In particular,
we augment a virtual node to the original graph bidirectionally connecting
to all real nodes (See Fig.~\ref{fig:Virtual-Column-Network} for
illustration). The corresponding virtual column hence performs two
functions: (i) integrating state information from all node columns,
and (ii) distributing the consensus graph-level information to all
node columns. The virtual column can optionally take graph-level information
as input (e.g., graph descriptors that are not encoded in the graph
structure).

With the virtual column, high-order and implicit dependencies are
distributed much faster, taking only 2 steps, independent of the graph
size. The virtual column and the node columns are computed as follow:

\begin{eqnarray*}
\hb_{0}^{t} & = & g\left(W_{0}\hb_{0}^{t-1}+\frac{1}{n}\sum_{i=1}^{n}U_{0}\hb_{i}^{t-1}\right)\\
\hb_{i}^{t} & = & g\left(W\hb_{i}^{t-1}+V\hb_{0}^{t-1}+\sum_{p}U_{p}\cb_{ip}^{t}\right)
\end{eqnarray*}

Let $\hb_{0}^{T}$ be the state of the virtual column at step $T$.
The iterative estimation can continue (a) with dimensional change,
which requires a projection onto a different state space, (b) without
input from other nodes. The later part of the column is essentially
a Highway Network \cite{srivastava2015training} with parameters typing
\cite{pham2016faster} (or a GRU if a reset gate is used \cite{li2016gated}).

\textbf{Remark}: In a way, it is similar to semi-restricted Boltzmann
machines, where the real columns and their connections handle short-range
dependencies, and the virtual column enables long-range dependencies.
The recurrent structure is akin to mean-field unrolled to $T$ steps.

\section{Experiments}

We demonstrate the effectiveness of our model against the baselines
on BioAssay activity prediction tasks and a code classification task.

\subsection{Experiment settings}

For all experiments with our proposed VCN, we use Highway network
for all layers and ReLU as activation function. Dropout \cite{srivastava2014dropout}
is applied before and after the column layers. Each dataset is divided
into training, validation and test sets. The validation set is used
for early stopping and tuning hyper-parameters. We set the number
of hidden layers by 10, the mini-batch by 100 and search for (i) hidden
dimensions of the virtual columns, (ii) the hidden dimensions of the
node columns, (iii) the learning rate $\eta$ (0.001, 0.002,...,0.005)
and (iv) optimizers: Adam \cite{kingma2014adam} or RMSprop. The training
starts at the learning rate $\eta$ and will be divided by 2 if the
model cannot find a better result on the validation set. Learning
is terminated after 4 times of halving or after 500 epochs. The best
setting is chosen by the validation set and the result of the test
set are reported.

Baselines are SVM, Random Forest and Gradient Boosting Machine. These
method reads input as vector representation of graphs. Sec.~\ref{subsec:Feature-NCI}
and Sec.~\ref{subsec:Feature-java} describe feature extraction methods
for the baselines. For BioAssay activity prediction, we add Neural
Fingerprint (NeuralFP) \cite{duvenaud2015convolutional} as a baseline.

\subsection{BioAssay Activity Prediction}

\subsubsection{Datasets}

The first set of experiments uses 3 largest NCI \textbf{BioAssay}
activity tests collected from the PubChem website \footnote{https://pubchem.ncbi.nlm.nih.gov/}:
Lung Cancer, Leukemia and Yeast Anticancer. Each BioAssay test contains
records of activities for chemical compounds. Each compound is represented
as a graph, where nodes are atoms and edges are bonds between them.
We chose the 2 most common activities for classification: ``active''
and ``inactive''. The statistics of data is reported in Table~\ref{tab:Summary-of-data}.
These datasets are unbalanced, therefore ``inactive'' compounds
are randomly removed so that each of Lung Cancer and Leukemia datasets
has 10,000 graphs and the Yeast Anticancer dataset has 25,000 graphs.

\begin{table}[h]
\centering{}%
\begin{tabular}{clcc}
\hline 
No. & Dataset & \# Active & \# Graph\tabularnewline
\hline 
1 & Lung Cancer & 3,026 & 38,588\tabularnewline
2 & Leukemia & 3,681 & 38,933\tabularnewline
3 & Yeast Anticancer & 10,090 & 86,130\tabularnewline
\hline 
\end{tabular}\caption{Summary of the three NCI BioAssay datasets. ``\# Graph'' is the
number of graphs and ``\# Active'' is the number of active graph
against a BioAssay test. \label{tab:Summary-of-data}}
\end{table}

\subsubsection{Feature extraction \label{subsec:Feature-NCI}}

We use RDKit toolkit for molecular feature extraction \footnote{http://www.rdkit.org/}.
RDKit computes fixed-dimensional feature vectors of molecules, which
is so-called circular fingerprint. These vectors are used as inputs
for the baselines. We set the dimension of the fingerprint features
by 1024.

For our model, we also use RDKit to extract the structure of molecules
and the atom features. An atom feature vector is the concatenation
of the one-hot vector of the atom and other features such as atom
degree and number of H atoms attached. We also make use of bond features
such as bond type and a binary value indicating if a bond in a ring.

\subsubsection{Results}

Table~\ref{tab:AUC-NCI} reports results, measured in AUC, on the
NCI datasets. The proposed Virtual Column Network (VCN) is competitive
against best feature engineering techniques (cicular fingerprint and
high-performing classifiers).

\begin{table}[h]
\begin{centering}
\begin{tabular}{lcccc}
\hline 
Method & Lung & Leukemia & Yeast & Average\tabularnewline
\hline 
FP+SVM & 85.1 & 82.1 & 77.3 & 81.5\tabularnewline
FP+RF & 85.2 & 82.1 & 76.5 & 81.3\tabularnewline
FP+GBM & 81.5 & 82.3 & 77.0 & 80.3\tabularnewline
NeuralFP & 85.5 & \textbf{84.5} & 79.5 & 83.2\tabularnewline
\textbf{VCN} &  \textbf{86.3} &  83.3 & \textbf{81.1} & \textbf{83.6}\tabularnewline
\hline 
\end{tabular}
\par\end{centering}
\caption{Area under the ROC curve (AUC) (\%) for NCI datasets. FP = Fingerprint;
RF = Random Forests; GBM = Gradient Boosting Machine; VCN = proposed
Virtual Column Network. \label{tab:AUC-NCI}}
\end{table}

\begin{figure}[h]
\centering{}\includegraphics[bb=30bp 0bp 525bp 400bp,clip,width=0.98\columnwidth]{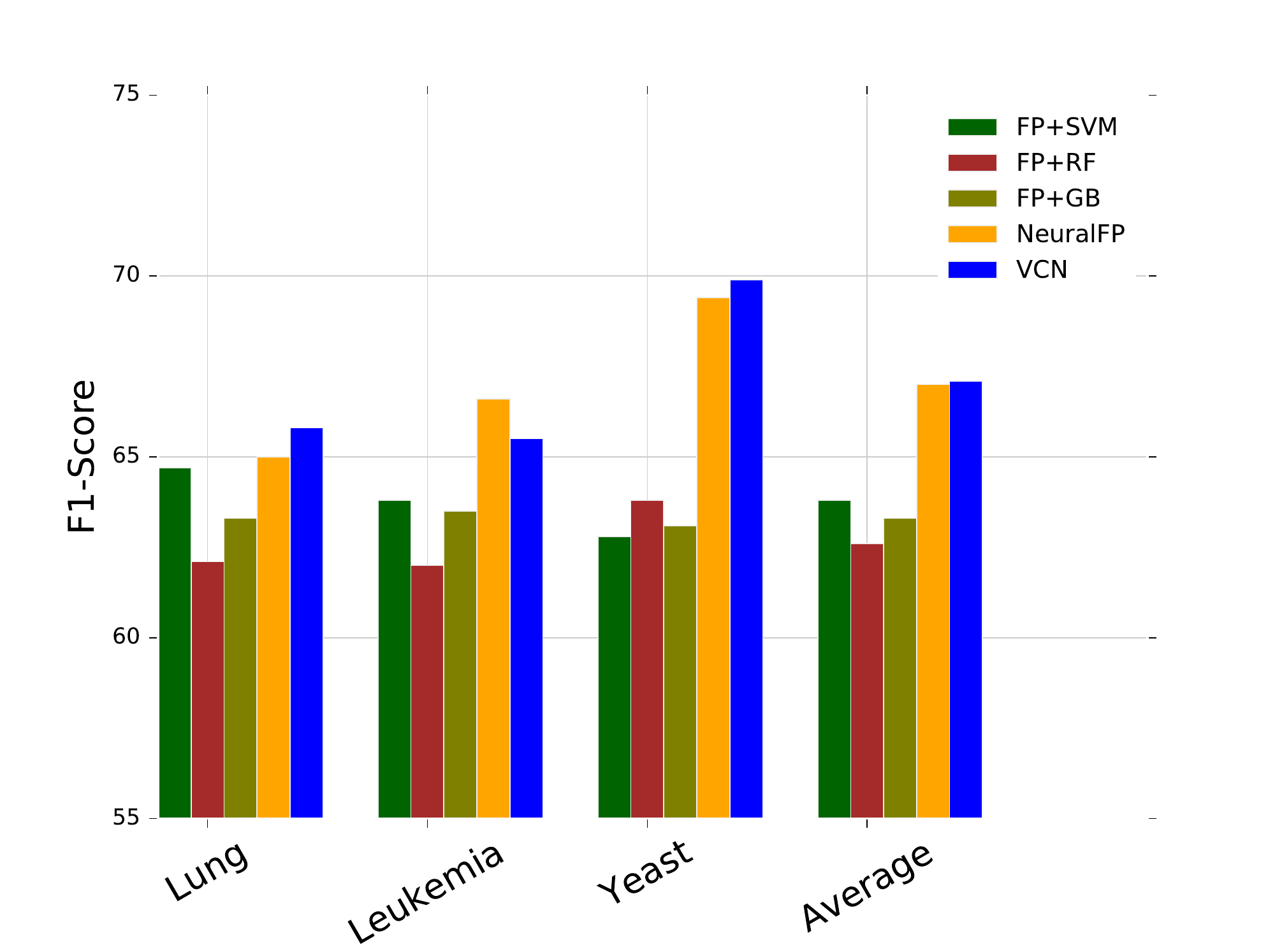}\caption{F1-score (\%) for NCI datasets. FP = Fingerprint; RF = Random Forests;
GBM = Gradient Boosting Machine; VCN = proposed Virtual Column Network.
Best view in color. \label{fig:F1-score}}
\end{figure}

Fig.~\ref{fig:F1-score} reports the F1-score on the NCI datasets.
On average, VCN beats all the baselines.

\subsection{Code Classification}

\subsubsection{Dataset}

The dataset contains 18 Java projects, each consists of a number of
source code files. Each source file is a Java class, which has a list
attribute declarations and a number of methods. A class is represented
as a graph, where graph-level features are the attribute declarations,
nodes are methods and edges are method call. The task is to predict
if a source file is vulnerable. The dataset is pre-processed by removing
all replicated files of different versions in the same projects. This
remains 2836 samples in total and 1020 positive ones.

\subsubsection{Feature extraction \label{subsec:Feature-java}}

Methods and attribute declarations of Java classes can be considered
as sequences of tokens and their representation can be learned through
language modeling using LSTM \cite{hochreiter1997long} with Noise
Contrastive Estimation (NCE) \cite{mnih2012fast}. The feature vector
of each sequence is the mean of all hidden states outputted by the
LSTM. After this step, each sequence is represented as a feature vector
of 128 units. For the baselines, the feature vector of each Java class
is the mean of feature vectors of all methods and attribute declarations.
For the VCN model, the feature vector of the attribute declarations
is the input for the virtual column.

\subsubsection{Results}

\begin{figure}

\begin{centering}
\includegraphics[bb=100bp 0bp 550bp 400bp,clip,width=0.8\columnwidth]{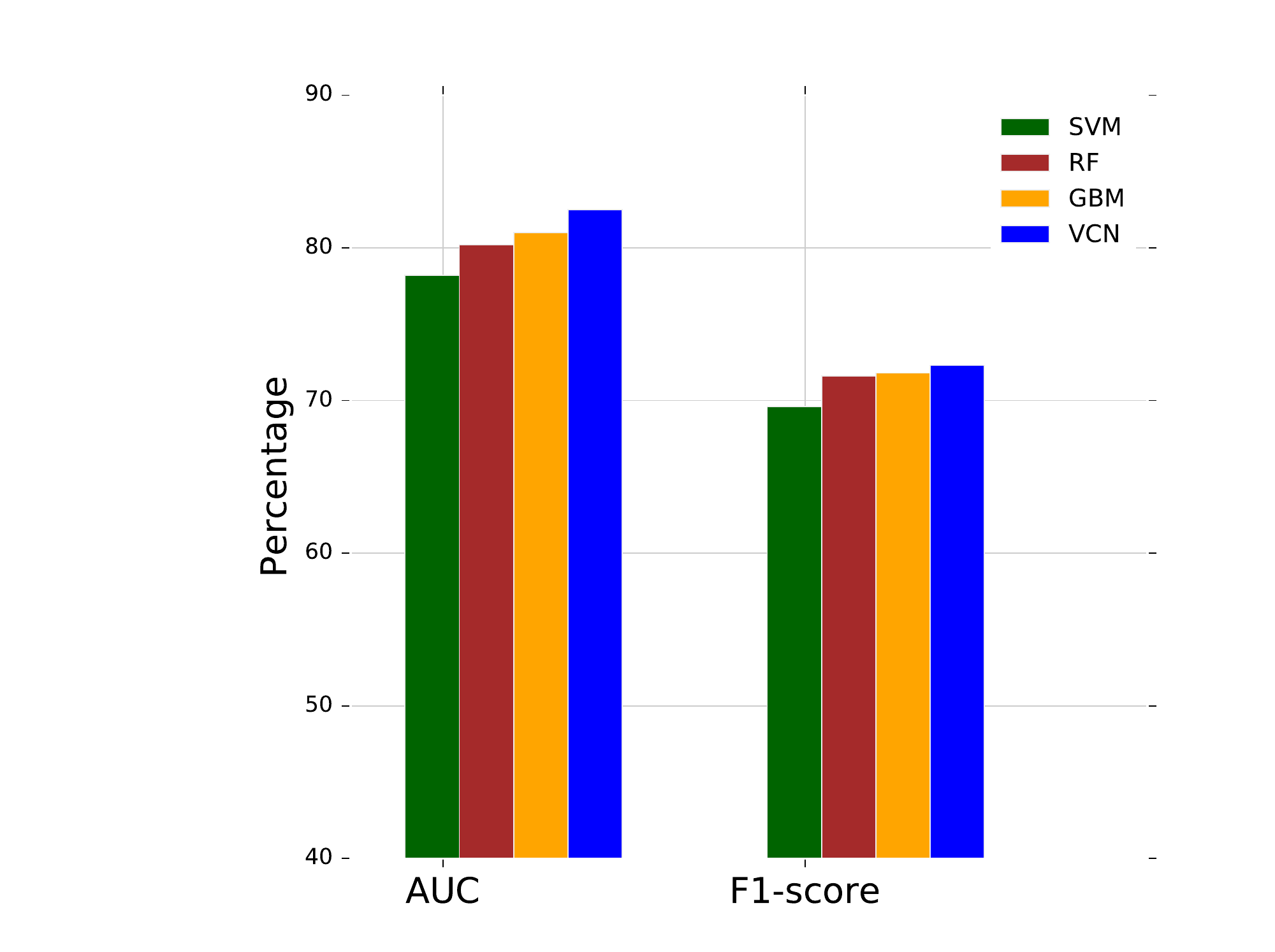}\caption{Performance on Code classification dataset, measured in AUC and F1-score
(\%).\label{fig:Performance-on-Code}}
\par\end{centering}
\end{figure}

Fig.~\ref{fig:Performance-on-Code} reports the performance on Code
classification task in AUC and F1-score. VCN outperforms all the baselines
on both measures.

\section{Related Work}

There has been a sizable rise of learning graph representation in
the past few years \cite{bronstein2016geometric,bruna2014spectral,henaff2015deep,johnson2017learning,li2016gated,niepert2016learning,schlichtkrull2017modeling}.
A number of works derive shallow embedding methods such as node2vec
and subgraph2vec, possibly inspired by the success of embedding in
linear-chain text (word2vec and paragraph2vec). Deep spectral methods
have been introduced for graphs of a given adjacency matrix \cite{bruna2014spectral},
whereas we allow arbitrary graph structures, one per graph. Several
other methods extend convolutional operations to irregular local neighborhoods
\cite{atwood2016diffusion,niepert2016learning,pham2017column}. Yet
recurrent nets are also employed along the random walk from a node
\cite{scarselli2009graph}. 

This paper is built upon our recent work, the Column Network (real
nodes only, designed for node classification) \cite{pham2017column},
and Column Bundle (no graphs, designed for multi-part data, where
part can be instance or view) \cite{pham2017one}. Like is predecessors,
it can be seen as an instance of learning as iterative estimation
\cite{greff2017highway}. 

Our application to chemical compound classification bears some similarity
to the work of \cite{duvenaud2015convolutional}, where graph embedding
is also collected from node embedding at each layer and refined iteratively
from the bottom to the top layers. However, our treatment is more
principled and more widely applicable to multi-typed edges.

\section{Discussion}

We have proposed a simple solution for learning representation of
a graph: adding a virtual node to the existing graph. The expanded
graph can then be passed through any node representation method, and
the representation of the virtual node is the graph's. The virtual
node, coupled with a recent node representation method known as Column
Network \cite{pham2017column}, results in a new graph classification
method called Virtual Column Network (VCN). We demonstrate the power
of the VCN on two tasks: (i) classification of bio-activity of chemical
compounds against a given cancer; (ii) detecting software vulnerability
from source code. Overall, the automatic representation learning is
more powerful than state-of-the art feature engineering.

There are rooms open for further investigations. First, we can use
multiple virtual nodes instead of just one.The graph is then embedded
into a matrix whose columns are vector representation of virtual nodes.
This will be beneficial in several ways. For multitask learning, each
virtual node will be used for a task and all tasks share the same
node representations. For big graphs with tight subgraph structures,
each virtual node can target a subgraph. Second, other node representation
architectures beside Column Networks are also applicable for deriving
graph representation, including Gated Graph Sequence Neural Network
\cite{li2016gated}, Graph Neural Network \cite{scarselli2009graph}
and diffusion-CNN \cite{atwood2016diffusion}.

\subsubsection*{Acknowledgments}

The paper is partly supported by the Samsung 2016 GRO Program titled
``Predicting hazardous software components using deep learning'',
and the Telstra-Deakin CoE in Big Data and Machine Learning.

\bibliographystyle{named}
\bibliography{../bibs/ME,../bibs/trang,../bibs/truyen}

\end{document}